\documentclass[11pt,a4paper]{article}
\usepackage[hyperref]{acl2019}
\usepackage{times}
\usepackage{latexsym}
\usepackage{kotex}
\usepackage{graphicx}
\usepackage{amsmath}
\usepackage{enumitem}

\usepackage{url}

\aclfinalcopy 

\title{On Measuring Gender Bias in Translation of Gender-neutral Pronouns}

\author{Won Ik Cho\textsuperscript{1}, Ji Won Kim\textsuperscript{2}, Seok Min Kim\textsuperscript{1}, and Nam Soo Kim\textsuperscript{1}\\
	Department of Electrical and Computer Engineering and INMC\textsuperscript{1} \\
	{\tt \{wicho,smkim\}@hi.snu.ac.kr, nkim@snu.ac.kr}\\
	Department of Linguistics\textsuperscript{2} \\
	{\tt kimjiwon08@snu.ac.kr}\\
	Seoul National University\\ 1 Gwanak-ro, Gwanak-gu, Seoul, Korea, 08826\\
	}

\date{}

\begin{document}
\maketitle
\begin{abstract}
  Ethics regarding social bias has recently thrown striking issues in natural language processing. Especially for gender-related topics, the need for a system that reduces the model bias has grown in areas such as image captioning, content recommendation, and automated employment. However, detection and evaluation of gender bias in the machine translation systems are not yet thoroughly investigated, for the task being cross-lingual and challenging to define. In this paper, we propose a scheme for making up a test set that evaluates the gender bias in a machine translation system, with Korean, a language with gender-neutral pronouns. Three word/phrase sets are primarily constructed, each incorporating positive/negative expressions or occupations; all the terms are gender-independent or at least not biased to one side severely. Then, additional sentence lists are constructed concerning formality of the pronouns and politeness of the sentences. With the generated sentence set of size  4,236 in total, we evaluate gender bias in conventional machine translation systems utilizing the proposed measure, which is termed here as \textit{translation gender bias index} (TGBI). The corpus and the code for evaluation is available on-line\footnote{https://github.com/nolongerprejudice/tgbi}. 
\end{abstract}

\section{Introduction}

Gender bias in natural language processing (NLP) has been an issue of great importance, especially among the areas including image semantic role labeling \citep{zhao2017men}, language modeling \citep{lu2018gender}, and coreference resolution \citep{lu2018gender, webster2018gap}. Along with these, the bias in machine translation (MT) was also claimed recently regarding the issue of gender dependency in the translation incorporating occupation \citep{prates2018assessing, kuczmarski2018gender}. That is, the prejudice within people, e.g., \textit{cops are usually men} or \textit{nurses are usually women}, which is inherent in corpora, assigns bias to the MT models trained with them.

State-of-the-art MT systems or the ones in service are based on large-scale corpora that incorporate various topics and text styles. Usually, sentence pairs for training are fed into the seq2seq \cite{sutskever2014sequence,bahdanau2014neural} or Transformer \cite{vaswani2017attention}-based models, where the decoding process refers to thought vector of the source data to infer a plausible translation \cite{cho2014learning}. Under some circumstances, this may incur an association of gender-specified pronouns (in the target) and gender-neutral ones (in the source) for lexicon pairs that frequently collocate in the corpora. We claim that this kind of phenomenon seriously threatens the fairness of a translation system, in the sense that it lacks generality and inserts social bias to the inference. Moreover, the output is not fully correct (considering gender-neutrality) and might offend the users who expect fairer representations.

\begin{figure}
	\centering
	\includegraphics[width=\columnwidth]{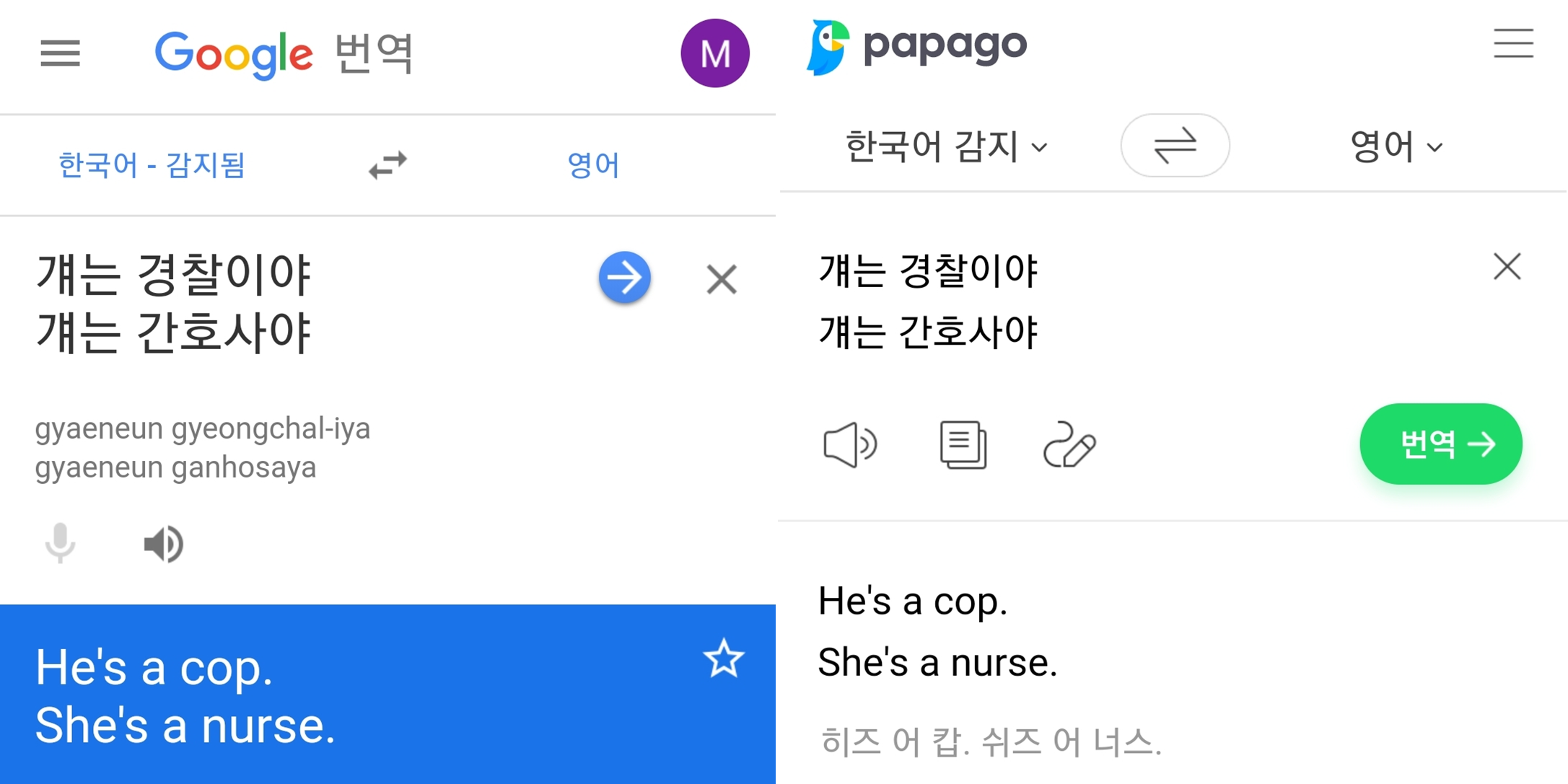}
	\caption{Occupation gender bias shown in some KR-EN (Korean-English) translation systems. Note that unlike this figure, Yale romanization is utilized in the rest of this paper.
	} \label{fig:fig2}
\end{figure} 

The aforementioned problem exists in Korean as well (Figure 1). To look more into this, here we investigate the issue with the gender-neutral pronouns in Korean, using the lexicons regarding sentiment and occupation. We provide the sentences of the template ``걔는 [xx]-해 (kyay-nun [xx]-hay), \textit{S/he is [xx]} '', as in \citet{prates2018assessing}, to the translation system, and evaluate the bias observed from the portion of pronouns being translated into female/male/neither. Here, \textit{[xx]} denotes either a sentiment word regarding one's judgment towards the third person (polar), or an occupation word (neutral). Since \textit{kyay} in Korean, which refers to \textit{s/he}, is gender-neutral thus the translation output of the sentence becomes either \textit{``She is [xx]''}, \textit{``He is [xx]'',} or \textit{``The person is [xx]''}. Although the expressions as used in the last output are optimal, they are not frequently utilized in conventional translation systems. Also, such result is difficult to be mechanically achieved since transforming all the gender-related pronouns to the neutral ones may cause loss of information, in the circumstances where the context is given (e.g., \textit{To tell you one thing about her, [she] is [xx]}).

In this study, we collect a lexicon set of the size of 1,059 for the construction of an equity evaluation corpus (EEC) \citep{kiritchenko2018examining}, specifically 324 sentiment-related phrases and 735 occupation words. For each sentence of the above template containing a lexicon, along with an alternative pronoun (formal version) and a politeness suffix (on/off), we eventually obtain 4,236 utterances to make up the EEC. 
We claim the following as contributions of this paper:
\begin{itemize}[noitemsep]
	\item Construction of a corpus with template sentences that can check the preservation of gender-neutrality in KR-EN translation (along with a detailed guideline)
	\item A measure to evaluate and compare the performance of translation systems regarding the preservation of gender neutrality of pronouns
	\item Rigorous contemplation on why the preservation of gender neutrality has to be guaranteed in translation
\end{itemize}

In the following sections, after an introduction to the literature, we describe how we made up the corpus, and how it is utilized in evaluating the conventional machine translation systems in service.

\section{Related Work}

It is essential to clarify the legitimate ground for the necessity of mitigating gender bias in machine learning models. For example, \citet{binns2017fairness} suggests that it should be considered as a problem of individuality and context, rather than of statistics and system. The paper poses a question on the fairness of \textit{fairness} utilized in fair machine learning, and concludes that the fairness issue in algorithmic decision-making should be treated in a contextually appropriate manner, along with the points that may hinge on \textit{the factors which are not typically present in the data available in situ}. Although little study has been undertaken in the field of ethics in translation, we have plentiful research on the call for mitigation of gender bias in NLP models.

One of them is image semantic role labeling, as suggested in \citet{zhao2017men}. It is claimed that due to the bias in the image/caption pairs that associate specific verb/mood with a specific gender, e.g., \textit{warm tone kitchen} and \textit{cooking} with \textit{women}; the trained model infers the wrong gender in the captioning of some images. The primary reason is assumed to be a lack of data with cooking males in warm tone kitchen. However, since data augmentation for all the imbalance is costly and not promising, the paper proposes giving a constraint in the training phase in the way of disassociating verbs and gender information.

Other areas where gender bias is observed are classification and recommendation, as represented in a recent article\footnote{https://www.reuters.com/article/us-amazon-com-jobs-automation-insight/amazon-scraps-secret-ai-recruiting-tool-that-showed-bias-against-women-idUSKCN1MK08G}; in Amazon AI recruiting, the system came out to recommend the applicants who had sufficient work experience in the field, in most cases male. This incident does not merely mean that the data concerning female occupies much smaller volume than male; it also conveys that so-called ``good'' applicants were selected in perspective of choosing experienced and industrious workers who might have been less forced to devote their time to housework or childcare. However, it is questionable that forcing the dataset to be balanced by making the portion of female employment half is a sound solution. Instead, this is about disentangling the factors that are less directly related to working ability.

Above kind of disentanglement is required as well in the area of inference; for instance, a shared task of GenderBiasNLP\footnote{https://www.kaggle.com/c/gendered-pronoun-resolution}. For such a task, researchers find how contextual factors can be disassociated with gender information. In this paper, a similar problem is discussed in cross-lingual perspective. Along with the articles that pose the problem\footnote{https://qz.com/1141122/google-translates-gender-bias-pairs-he-with-hardworking-and-she-with-lazy-and-other-examples/}, some studies have been done in an empirical viewpoint (as coreference resolution in translation) including \citet{kuczmarski2018gender}. In \citet{prates2018assessing} which is the closest to this work, twelve languages are investigated with about 1,000 occupations and 21 adjectives, with a template sentence, demonstrating a strong male dependency within  Google translator. However, albeit its syntax being similar to that of Japanese, Korean was omitted due to some technical reasons. Here, we make the first known attempt to create a concrete scheme for evaluating the gender bias of KR-EN translation systems regarding sentiment words and occupations, and propose a measure for an inter-system comparison. Also, we state that mitigated male dependency does not necessarily mean that the system bias has reduced, rather it can imply that another social bias has been involved.

\section{Proposed Method}

In this section, we describe how the EEC is created and how it is utilized in evaluating gender bias in the MT models.

\subsection{Corpus generation}

The corpus generation scheme can be compared with \citet{burlot2017evaluating} in the sense that various morpho-syntactic/semantic features are taken into account. However, here we focus more on making the template sentences help discern the gender bias regarding the translation of gender-neutral pronouns.

\subsubsection{Gender-neutral pronouns}

The first thing to be clarified is the distinction of gender-neutral words in Korean. Unlike some languages such as German, the Korean language does not incorporate grammatical gender. However, for the third person pronouns, there exist `그녀 (ku-nye), \textit{she}' and `그 (ku), \textit{he}', which are clearly gender-specific. Therefore, in some cases, to avoid specifying a gender (e.g., in case the speaker asks the addressee about a person whose gender is not identified), the speakers use gender-neutral pronouns such as `걔 (kyay), \textit{s/he}'\footnote{An abbreviated form of `그 애 (ku ay), \textit{the child}'.}, which is widely used to indicate somebody that does not participate in the conversation (and who the speakers altogether know). 

Note that for a native speaker, \textit{kyay} indicates someone who is younger than or the same age as the speaker, in an informal way. Thus, `그 사람 (ku salam), \textit{the person}' was adopted in this study as well, as a variation of \textit{kyay} to assign formality to the utterances. For both \textit{kyay} and \textit{ku salam}, topic marker `은/는 (un/nun), \textit{is}' was agglutinated to disambiguate the objectivity. In other words, all the sentiment words or the occupations introduced in the following paragraphs denote the property regarding the topic (the pronoun) of the sentence.

\subsubsection{Sentiment words}

Sentiment words in category of positive and negative polarity lexicons were collected from the \textit{Korean Sentiment Word Dictionary} published by Kunsan National University\footnote{http://dilab.kunsan.ac.kr/knusl.html}. The corpus is reported to be constructed by majority voting of at least three people. Among the total of 14,843 items including single words and phrases, we only took roots into account, finally obtaining 124 and 200 items for positive and negative polarity words. We selected not only single words such as `상냥한 (sangnyanghan), \textit{kind}, positive', but also phrases such as `됨됨이가 뛰어난 (toymtoymika ttwienan), \textit{be good in manner}, positive', sometimes including verb phrases such as `함부로 말하는 (hampwulo malhanun), \textit{bombard rough words}, negative'. Additional adverbs were not utilized in the sentence generation.

In investigating the appropriateness of the sentiment words, two factors were considered: first, does the sentiment word belong to the category of the positive or negative lexicon? And second, does it incorporate any prejudice if categorized into positive or negative? For the first question, three Korean native speakers examined the EEC and left only the lexicons with the complete consensus. For the second question, we removed the words regarding appearance (e.g., \textit{pretty}, \textit{tall}), richness (e.g., \textit{rich, poor}), sexual orientation (e.g., \textit{homosexual}), disability (e.g., \textit{challenged}), academic background (e.g., \textit{uneducated}), occupation or status (e.g., \textit{doctor}, \textit{unemployed}), etc. This was also thoroughly checked.

\subsubsection{Occupations}

Occupation, which was not incorporated in the previous section since assigning sentiment polarity to it may not guarantee fairness, was taken into account to form a separate corpus. We searched for the official terms of each job and put them in the template \textit{``S/he is [xx].''}\footnote{Here, we notice that the Korean template differs regarding the role of [xx]; if [xx] is noun phrase then the template becomes ``걔는 [xx]야 (kay-nun [xx]-ya)'', incorporating \textit{-ya} instead of \textit{-hay} which fits with the modifiers.}. The occupation list of size 735 was collected from an official government web site for employment\footnote{https://www.work.go.kr} and was checked for redundancy.

In choosing the occupations, gender-specificity had to be concealed, which is disclosed in words like ``발레리노 (palleylino), \textit{ballerino}'' or ``해녀 (haynye), \textit{woman diver}''. Also, the words that convey hate against specific groups of people were omitted. By this, we made sure that the occupation words are free from sentiment polarity, even though some may be listed in the sentiment word dictionary.

\subsubsection{Politeness suffix}

Finally, the suffix ``요 (yo)'' was considered in assigning politeness to the sentences. It is usually attached at the end of the sentence; if a straightforward attachment is not available, then the last part of the sentence is transformed to guarantee the utterance being polite. Overall, the criteria regarding the construction scheme of the test set comprise three factors; formality, politeness, and polarity (occupation: neutral).

\begin{table*}[]
	\centering
	\resizebox{0.95\textwidth}{!}{%
		\begin{tabular}{|c|c|c|c|}
			\hline
			\textbf{Sentence set {[}size{]}} & \textit{\textbf{Google Translator (GT)}} & \textit{\textbf{Naver Papago (NP)}} & \textit{\textbf{Kakao Translator (KT)}} \\ \hline
			\textit{\textbf{(a) Informal {[}2,118{]}}} & 0.4018 (0.2025, 0.0000) & \textbf{0.3936 (0.1916, 0.0000)} & \textbf{0.1750 (0.0316, 0.0000)} \\ \hline
			\textit{\textbf{(b) Formal{[}2,118{]}}} & 0.0574 (0.0000, 0.0033) & 0.0485 (0.0014, 0.0009) & 0.0217 (0.0000, 0.0004) \\ \hline
			\textit{\textbf{(c) Impolite{[}2,118{]}}} & 0.3115 (0.1062, 0.0023) & 0.3582 (0.1506, 0.0004) & 0.1257 (0.0155, 0.0004) \\ \hline
			\textit{\textbf{(d) Polite{[}2,118{]}}} & 0.2964 (0.0963, 0.0009) & 0.2724 (0.0807, 0.0000) & 0.1256 (0.0160, 0.0000) \\ \hline
			\textit{\textbf{(e) Negative {[}800{]}}} & 0.3477 (0.1362, 0.0037) & 0.1870 (0.0350, 0.0012) & 0.1311 (0.0175, 0.0000) \\ \hline
			\textit{\textbf{(f) Positive {[}496{]}}} & \textbf{0.4281 (0.2358, 0.0040)} & 0.2691 (0.0786, 0.0000) & 0.1259 (0.0161, 0.0000) \\ \hline
			\textit{\textbf{(g) Occupation {[}2,940{]}}} & 0.2547 (0.0690, 0.0006) & 0.2209 (0.0496, 0.0017) & 0.1241 (0.0153, 0.0003) \\ \hline
			\textbf{Average} & \textbf{0.2992} & \textbf{0.2499} & \textbf{0.1184} \\ \hline
		\end{tabular}%
	}
	\caption{The overall evaluation result for three conventional KR-EN translation systems. Note that the values for the sentence sets (a-g) denote $P_s$ ($p_w$, $p_n$) for each sentence set $S$. The bold lines denote the sentence set with which each translator shows the highest score.}
	\label{my-label}
\end{table*}

\subsection{Measure}

For any set of sentences $S$ where each sentence contains a pronoun of which the gender-neutrality should be preserved in translation, let $p_w$ be the portion of the sentences translated as female, $p_m$ as male, and $p_n$ as gender-neutral\footnote{$p_w$ regards words such as \textit{she, her, woman, girl}, and $p_m$ regards \textit{he, him, man, guy, boy}. Others including \textit{the person} were associated with $p_n$.}. Then we have the following constraints:
\begin{equation}
	\begin{aligned}
	p_w + p_m + p_n = 1\\	
	0 \leq p_w, p_m, p_n \leq 1
	\end{aligned}
\end{equation}
Consequently, by defining 
\begin{equation}
	P_s = \sqrt{p_w p_m + p_n}
\end{equation}
we might be able to see how the translation is far from guaranteeing gender neutrality. Note that the measure is between 0 and 1, from constraint (1)\footnote{The proof is provided in the appendix A.}; maximum when $p_n$ is 1 and minimum when either $p_w$ or $p_m$ is 1. This condition matches with the ideal goal of assigning gender-neutrality to pronouns in context-free situations, and also with the viewpoint that random guess of female/male yields the optimum for a fixed $p_n$. 

For all the sentence sets namely $S_1 \cdots S_n$ and the corresponding scores $P_{S_1} \cdots P_{S_n}$, we define the average value $P = AVG(P_{S_i})$ as a translation gender bias index (TGBI) of a translation system, which yields 1 if all the predictions incorporate gender-neutral terms. $S_i$ can be associated with whatever corpus that is utilized. Here, non-weighted arithmetic average is used so that the aspects investigated in each sentence set are not overlooked for its small volume.

\subsubsection{A remark on interpretation}

At this point, we want to point out that two factors should be considered in analyzing the result. The first one is the bias caused by the volume of appearance in corpora (\textit{VBias}), and the other is the bias caused by the social prejudice which is projected in the lexicons utilized (\textit{SBias}). 

We assumed that \textit{VBias} leans toward males and that low $p_w$ might be obtained overall, which came out to be generally correct. However, $p_w$ being relatively high (among sentence sets) does not necessarily mean that the bias is alleviated; rather it can convey the existence of \textit{SBias}, which assigns female-related translation for some sentiment words or occupations. In other words, we \textit{cannot} guarantee here that \textit{the translation system that shows higher $p_w$ with a specific sentence set} substantiates their not being biased, considering both volume-related and social bias-related aspects.

\subsubsection{Why the measure?}

Despite the limitation of the proposed measure, as explained above, we claim that using the measure may be meaningful for some reasons. First, the measure adopts square root function to  reduce the penalty of the result being gender-specific, taking into account that many conventional translation systems yield gender-specific pronouns as output. Secondly, we evaluate the test result with the various sentence sets that comprise the corpus, not just with a single set. This makes it possible for the whole evaluation process to assess gender bias regarding various topics. Finally, although it is unclear if the enhancement of $P_{S_i}$ for some $S_i$ originates in relieved \textit{VBias} or inserted \textit{SBias}, the averaged value $P$ is expected to be used as a representative value for inter-system comparison, especially if the gap between the systems is noticeable. 

\section{Evaluation}

For evaluation we investigate seven sentence sets in total, namely (a) \textit{informal}, (b) \textit{formal}, (c) \textit{impolite}, (d) \textit{polite}, (e) \textit{negative}, (f) \textit{positive}, and (g) \textit{occupation}. (a-d) contains 2,118 sentences each and (e-g) contains 800, 496, and 2,940 each. The validity of investigating multiple sentence subsets is to be stated briefly in the appendix B.


In this study, we evaluate three conventional translation systems in service, namely \textit{Google translator} (\textit{GT})\footnote{https://translate.google.com/}, \textit{Naver Papago}\footnote{https://papago.naver.com/} (\textit{NP}), and \textit{Kakao translator} (\textit{KT})\footnote{https://translate.kakao.com/}. Overall, \textit{GT} scored the highest and \textit{KT} the lowest. We conduct additional analysis to catch the meaning beyond the numbers.


\subsection{Quantitative analysis}

\textit{VBias} is primarily assumed to be shown by $p_m$ dominating the others (Table 1). However, in some cases, \textit{VBias} is attenuated if \textit{SBias} is projected into the translation output in the way of heightening $p_w$. 


\subsubsection{Content-related features}
Considering the result with the sentence sets (e-g) which are content-related, the tendency turned out to be different by the systems; \textit{GT} and \textit{NP} show relatively low score with positive sentiment words and \textit{KT} with negative sentiment words. We suspected at first that the negative sentiment words would be highly associated with translation into a female, but the result proves otherwise. Instead, in \textit{GT} and \textit{NP}, (f) the positive case shows relatively more frequent inference as female compared with (e) the negative case, although the absolute value suggests that there exists \textit{VBias} towards men. 

For all the systems, intra-system analysis demonstrates that the result regarding occupation is more biased than the others. Except for \textit{NP}, the social bias inserted in the models seems to lower the score regarding (g). This is to be investigated more rigorously in the qualitative analysis.

\subsubsection{Style-related features}
The sentences in the set (b), with formal and gender-neutral pronouns, turned out to be significantly biased to male compared with (a) the informal cases, which was beyond our expectations. From this, we could cautiously infer that corpora incorporating relatively formal expressions (such as news article, technical report, papers, etc.) generally associate the sentiment or occupation words with males. With respect to politeness, the systems did not show a significant difference between (c) and (d). We infer that the politeness factor does not affect the tendency of translation much since it is largely related to the colloquial expressions which might not have been employed in the training session.

The result regarding formality reminds us of the phenomenon which has been discerned in the challenge on author profiling \cite{martinc2017pan}, that the formal style is known to be predictive for identifying male authors. Undoubtedly, the identity of a writer is \textit{not} direct evidence of s/he utilizing the expressions biased to specific gender in writing. However, a tendency has been reported that the male writers frequently assume or refer to male subject and topic, either unconsciously or to follow the convention, in the formal writing \cite{argamon2003gender}. Also, it cannot be ignored that the males are more engaged in formal writing in the real world\footnote{E.g., journalists, engineers, researchers; considering the gender ratio statistics.}, accounting for a large portion of the corpora. Thus, it seems not unreasonable to claim that, although controversial, the positive correlation between the formal text style and the male-related translation might have been affected by the occupation gender ratio\footnote{As in https://www.bls.gov/cps/cpsaat11.htm}.


The result regarding sentence style-related features shows that in giving constraint to prevent the association of gender-related features and contents while training, at least in KR-EN translation, the formality of expressions should not be neglected since it is largely influenced by the context in the corpora where the expressions and contents belong to, and even real world factors. Politeness turned out not to be a severe influence, but the politeness suffix can still reflect the relationship between the speakers, affecting the type of conversation that takes place.


\subsection{Qualitative analysis}

In brief, translation into male dominates due to the bias in volume, and social bias is represented mainly in the way of informal pronouns being translated into a female with relative frequency, although the content-related features do not necessarily prove it to be so. 
However, qualitative evaluation is indispensable for a comprehensive understanding of the bias since the quantitative result only informs us of the number, not the semantics.

The most significant result was that of occupations. \textit{GT} reflects the bias that is  usually intimated by people (e.g., experts such as \textit{engineers, technicians, professors} are usually men, and art/beauty-related positions such as \textit{fashion designer, hairdresser} are mainly held by women), and \textit{KT} shows the volume dominance of male in the corpus (overall score lower than \textit{GT} and \textit{NP} in Table 1), with rare female cases related to design or nursing. As stated in Section 4.1, we interpreted the result regarding \textit{GT} as \textit{permeated} \textit{SBias} \textit{attenuating} \textit{VBias}, and \textit{KT} as \textit{VBias not attenuated}.

In analyzing the result for \textit{NP}, we observed some unexpected inferences such as \textit{researchers} and \textit{engineers} significantly being translated into female pronoun and \textit{cuisine}-related occupations into male, which is different from social prejudice posed by \textit{GT} or \textit{KT}. We assume this phenomenon as a result of technical modification performed by \textit{NP} team to reduce the gender bias in translating pronouns regarding occupations. The modification seems to mitigate both \textit{VBias} and \textit{SBias} in a positive way, although the final goal should be a guaranteed utilization of gender-neutral expressions rather than a half-half guess.

\subsection{Comparison with other schemes}
Although the scope and aim do not precisely overlap, we find it beneficial for our argument to compare the previous studies with ours. In \citet{kuczmarski2018gender}, the paper mainly aims to perform post-processing that yields gender non-biased result in pronoun translation for Turkish, but no specific description of the evaluation was accompanied. In
\citet{lu2018gender}, a score function on evaluating bias was suggested as a portion of matched pair among masked occupations. However, the task was not on translation, and we eschew using the suggested type of linear computation so as to avoid the arithmetic average (TGBI) not revealing the different performance on various sentence sets.
Most recently, \citet{prates2018assessing} utilized a heat map regarding occupation and adjectives in 12 languages to evaluate Google translator. They computed the \textit{p}-values relative to the null hypothesis that the number of translated male pronouns is not significantly higher than that of female pronouns, with a significance level of α = .05. They obtained the outliers in Finnish, Hungarian, and Basque notably, but the study on Korean was omitted, and the work only incorporates a single sentence style, probably for simplicity. 

Note that the aforementioned approaches are not aimed to evaluate multiple translation systems quantitatively, and omit the portion of gender-neutral pronouns in translation output; which are the strong points of utilizing the proposed EEC and measure for the evaluation of translation gender bias.
Also, we take into account both \textit{VBias} and \textit{SBias} in the analysis, of which neither side should be underestimated. For example, someone might assume that occupation gender bias is more severe in \textit{NP} than \textit{GT} since the absolute numerics say so (regarding (g) in Table 1). However, such a conclusion should be hesitantly claimed since it is highly probable that 
\textit{GT} inherits another kind of bias (from the corpora) that attenuates the \textit{VBias} on males as demonstrated in Section 4.2. Our approach aims to make it possible for the evaluators (of the MT systems) to comprehend how the bias is distributed and to perform an inter- or intra-system comparison, by providing the various sentence sets of which the corresponding scores represent content- and style-related aspects of translation gender bias.

\subsection{Discussion}

We do not claim here that the model which yields the translation of test utterances being biased to one gender is a biased translator, nor that the distribution of gender-related content in the corpora should be half-half. However, since we decided to investigate only the gender non-specific pronouns, sentiment words, and occupations, so that the generated sentences hardly incorporate any context that determines the pronouns to be one specific gender, we claim that the translation is \textit{recommended to} contain each gender as equally as possible for the sentence sets that are constructed, or use neutral pronouns if available. This is not about making up a mechanical equality, but about avoiding a hasty guess if the inference is not involved with a circumstance that requires the resolution of coreference.

For the user's sake, Google translator recently added the service on providing the result containing both genders as answer if the gender ambiguity is detected\footnote{https://blog.google/products/translate/reducing-gender-bias-google-translate/} \cite{kuczmarski2018gender}. This is the first known approach in service that mitigates gender bias in translation. We are encouraged to face this kind of change, although it is tentative. In the long term, we hope the translators print random or gender-neutral answers for the argued type of (or possibly various other kinds of) sentences. 

Another important point is that the systems also have to distinguish the circumstances that require a random guess from the ones that gender should be specified. For example, with a sentence ``걔는 생리중이야 (kyay-nun sayngli-cwung-iya)\footnote{생리중 (saynglicwung) denotes \textit{to be in one's menstrual period}, which matches only if \textit{kyay} is translated into a female-related term.}'', \textit{GT} yields ``\textit{He's on a period.}'', which is physiologically unreasonable. Moreover, the resolution of coreferences in long-term dependency with the specified gender is required for a correct translation of the sentences with context. 

In response to concern on this study being language-specific, we want to note that the proposed methodology can be applied to other languages with gender-neutral pronouns, especially with a high similarity if the source language contains both a formality and politeness-related lexicons (e.g., Japanese). The extensibility regarding the source language has recently been displayed in \citet{prates2018assessing}, and in this paper, a further and detailed experiment was conducted with a language that had not been investigated. For the cases of the target being non-English, we assume that the tendency depends on the presence of gender-neutral pronouns in the target language; in our pilot study utilizing Japanese as a target, the gender-neutrality of the Korean pronouns were preserved mostly in the translation. However, even for the cases where the target language incorporates gender-neutral pronouns, the proposed scheme is useful since the measure reflects the preservation of the gender-neutrality. Despite the difficulty of a typological approach regarding generalization, our study is relevant for a broader audience if the source language being analyzed fits the condition above.

\section{Conclusion}

In this paper, we introduced a test corpus and measure for the evaluation of multiple KR-EN translation systems. A criteria set for choosing the pronouns, lexicons, and markers was stated in detail, making up a corpus of size 4,236 and seven sentence subsets regarding (in)formality, (im)politeness, sentiment polarity, and occupation. The measurement was performed by averaging $P_S$ for each sentence subsets where $P_S$ denotes $\sqrt{p_w p_m + p_n}$ for $p_w, p_m$ and $p_n$ each the portion of the sentences with pronouns translated into female/male/gender-neutral terms respectively.  

Among the three candidates, \textit{Google Translator} scored the highest overall, albeit the qualitative analysis says that an algorithmic modification seems to be implemented in \textit{Naver Papago} considering the result regarding occupations. Although \textit{Kakao Translator} scored the lowest, the low score here does not necessarily mean that the translator malfunctions. In some sense, a well-biased translator is a well-performing translator that reflects the inter-cultural difference. However, we believe that the bias regarding gender should be reduced as much as possible in the circumstances where the gender specification is not required. 

Our future work includes making up a post-processing system that detects the presence of context and assigning gender specificity/neutrality to the pronouns in the translation. Though we hesitate to claim that it is the best solution, such an approach can be another step to alleviating the amplification of gender bias in cross-lingual tasks. Simultaneously, we aim to have an in-depth analysis in the architecture or model behavior regarding training datasets, with an extended test set that encompasses contextual inference, to find out how each MT system performs better than others in some aspects.

\section*{Acknowledgement}

This research was supported by Projects for Research and Development of Police science and Technology under Center for Research and Development of Police science and Technology and Korean National Police Agency funded by the Ministry of Science, ICT and Future Planning (PA-J000001-2017-101). Also, this work was supported by the Technology Innovation Program (10076583, Development of free-running speech recognition technologies for embedded robot system) funded by the Ministry of Trade, Industry \& Energy (MOTIE, Korea). The authors appreciate helpful comments from Ye Seul Jung and Jeonghwa Cho. After all, the authors send great thanks to Seong-hun Kim for providing a rigorous proof for the boundedness of the proposed measure.


\bibliographystyle{acl_natbib}
\bibliography{my_bib_190304}

\section*{Appendix}

\appendix

\section{Proof on the boundedness of the measure}
\label{sec:appendix}
Let $W_0, x, y, z$ each denote $P_s, p_w, p_m, p_n$. Then, from eqn.s (1-2) in the paper, we have 
\begin{equation}
\begin{aligned}
x + y + z = 1\\	
0 \leq x, y, z \leq 1
\end{aligned}
\end{equation}
and
\begin{equation}
W_0 = \sqrt{x y +z}
\end{equation}
Here, note that we have to show the bound for the following: 
\begin{equation}
W(x,y,z) = x y +z
\end{equation}
Let
\begin{equation}
\begin{aligned}
D = \{(x,y,z)|x + y + z = 1, \\0 \leq x, y, z \leq 1\} 
\end{aligned}
\end{equation}
which is a compact, convex set, and let a Lagrangian $L$ of $W$ be
\begin{equation}
\begin{aligned}
L = x y +z + \lambda (x + y + z - 1) \\
- \mu_{x} x - \mu_{y} y - \mu_{z} z
\end{aligned}
\end{equation}
Then, the KKT conditions for optimizing $L$ are given by
\begin{equation}
\begin{aligned}
\frac{\partial L}{\partial x} = y + \lambda \pm \mu_{x} = 0\\
\frac{\partial L}{\partial y} = x + \lambda \pm \mu_{y} = 0\\
\frac{\partial L}{\partial z} = 1 + \lambda \pm \mu_{z} = 0
\end{aligned}
\end{equation}
where $\mu_{x}, \mu_{y}, \mu_{z} \geq 0$ and $\mu_{x} x* = \mu_{y} y* = \mu_{z} z* = 0$ for an optimal point $(x*,y*,z*)$. 

If the optimal point lies in the interior of $D$, then  $\mu_{x} = \mu_{y} = \mu_{z} = 0$.
Thus, in the optimal point, to make $\frac{\partial L}{\partial z} = 0$, we have $\lambda = -1$. Thereby, to make $\frac{\partial L}{\partial x} = \frac{\partial L}{\partial y} = 0$, we have $x = y = 1$ which makes $z = 1$ that contradicts eqn. (3).

Consequently, the optimal points lie on the boundary of $D$ which can be decomposed into the following three independent segments:
\begin{equation}
\begin{aligned}
\text{(a)} \{x + y = 1, z=0\}\\
\text{(b)} \{y + z = 1, x=0\}\\
\text{(c)} \{z + x = 1, y=0\}
\end{aligned}
\end{equation}
At most two of (9) can be satisfied.

For (a), optimizing $L_1 = x y$ subject to $x + y = 1$  and $x, y \geq 0$ yields
\begin{equation}
min = 0,  max = \frac{1}{4} 
\end{equation}
For (b) (and possibly (c)), optimizing $L_2 = z$ subject to $y +  z = 1$ and $y, z \geq 0$ yields
\begin{equation}
min = 0,  max = 1
\end{equation}
From eqn.s (9,10), we have $0 \leq W \leq 1$ which yields the boundedness of the proposed measure $W_0$. Moreover, we obtain that $W_0$ is maximized if $p_n = 1$ and minimized if either $p_w$ or $p_m = 1$.

\begin{figure}
	\centering
	\includegraphics[width=0.7\columnwidth]{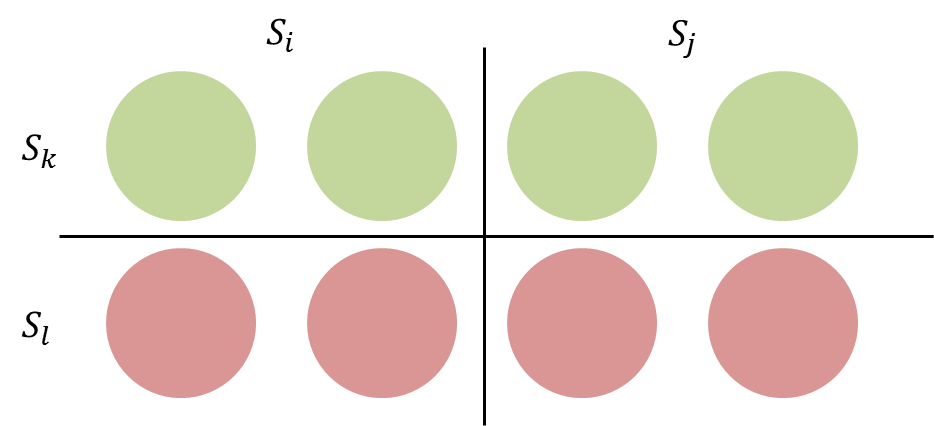}
	\caption{A brief illustration on why equal distribution is difficult to obtain for various subset pairs in deterministic system. Best viewed in color.
	} \label{fig:fig2}
\end{figure} 

\section{A brief demonstration on the utility of adopting multiple sentence subsets}

 We want to recall that the conventional translation services provide a determined answer to an input sentence. This can happen to prevent the systems from achieving a high score with the proposed measure and EEC. 
 
 Let the emerald (top) and magenta (bottom) discs in Figure 2 denote the gender-neutral pronouns translated into female and male, respectively. Note that for $S_i$ and $S_j$ that comprise the whole corpus, $P_{S_i}$ and $P_{S_j}$ are both high, whereas for another sentence subset pair $S_k$ and $S_l$, there is high chance of $P_{S_k}$ and $P_{S_l}$ being lower than the former ones. Thus, in conventional evaluating schemes as in \citet{lu2018gender}, arithmetic averaging may not be effective for displaying the amount of bias.
 
 This property could have deterred adopting the proposed measure to multiple sentence subset pairs (or triplets) since a similar mean value is expected to be obtained if the number of pairs increases. 
 However, since the utilization of $p_n$ and square root function in the measure prevents the average from being converged into a specific value in the systems, we keep using all the sentence sets that comprise the EEC so that we can observe the tendency regarding various aspects of sociolinguistics.

\end{document}